\documentclass[runningheads]{llncs}
\usepackage[T1]{fontenc}
\usepackage{array}
\usepackage{booktabs}
\usepackage{makecell}
\usepackage{multicol,multirow,booktabs}
\usepackage{subcaption}
\usepackage{amsmath}
\usepackage{graphicx,verbatim}
\usepackage[most]{tcolorbox}
\usepackage{xcolor}
\usepackage{listings}
\tcbset{
  markdownbox/.style={
    colback=gray!10,
    colframe=gray!60,
    arc=2mm,
    boxrule=0.4pt,
    fonttitle=\bfseries,
    left=2mm, right=2mm, top=1mm, bottom=1mm,
    enhanced
  }
}
\begin{document}
\title{Scan-do Attitude: Towards Autonomous CT Protocol Management using a Large Language Model Agent}
\titlerunning{Towards Autonomous CT Protocol Management}
%
%
%
\author{Xingjian Kang\inst{1}$^{,\dagger}$ 
\and
Linda Vorberg\inst{1,2} 
\and
Andreas Maier\inst{1}
\and
Alexander Katzmann\inst{2}$^{,*}$ 
\and
Oliver Taubmann\inst{2}$^{,*}$ }

\institute{Pattern Recognition Lab, Friedrich-Alexander Universität Erlangen-Nürnberg, Erlangen, Germany\\
\email{xingjian.kang@fau.de}\\
\and
Computed Tomography, Siemens Healthineers, Forchheim, Germany
\email{\{alexander.katzmann,oliver.taubmann\}@siemens-healthineers.com}}
\authorrunning{X.~Kang et al.}
\maketitle              
%
\begingroup
\renewcommand\thefootnote{}\footnotetext{$^{\dagger}$ Corresponding author}
\renewcommand\thefootnote{}\footnotetext{$^{*}$ These authors contributed equally to this work.}
\addtocounter{footnote}{-1}
\endgroup
\begin{abstract}
Managing scan protocols in Computed Tomography (CT), which includes adjusting acquisition parameters or configuring reconstructions, as well as selecting postprocessing tools in a patient-specific manner, is time-consuming and requires clinical as well as technical expertise. At the same time, we observe an increasing shortage of skilled workforce in radiology. To address this issue, a Large Language Model (LLM)-based agent framework is proposed to assist with the interpretation and execution of protocol configuration requests given in natural language or a structured, device-independent format, aiming to improve the workflow efficiency and reduce technologists' workload. The agent combines in-context-learning, instruction-following, and structured tool-calling abilities to identify relevant protocol elements and apply accurate modifications. In a systematic evaluation, experimental results indicate that the agent can effectively retrieve protocol components, generate device-compatible protocol definition files, and faithfully implement user requests. 
Despite demonstrating feasibility in principle, the approach faces limitations regarding syntactic and semantic validity due to lack of a unified device API, and challenges with ambiguous or complex requests. 
In summary, the findings show a clear path towards LLM-based agents for supporting scan protocol management in CT imaging. 

\end{abstract}
\section{Introduction}
In CT imaging, design and implementation of scan protocols as well as the dynamic adaptation of individual steps, including acquisition parameters, reconstruction settings and automated postprocessing, to patient-specific needs during the actual exam, require both domain-specific clinical knowledge and system-level technical expertise. These tasks are not only resource-intensive -- also due to the fact that different scanner models have different capabilities and different UI concepts -- but are increasingly difficult to manage amid a growing shortage of qualified staff in radiology~\cite{alexander2022mandating}.

Recent advancements in LLMs demonstrate remarkable potential in supporting daily work in radiology~\cite{d2024large}. Extensive research has also shown that LLMs can be guided to edit structured data: Hegselmann et al.~\cite{hegselmann2023tabllm} and Jiang et al.~\cite{jiang2023structgpt} both flattened structured data into LLM-interpretable text strings to improve LLM performance on structured data classification and reasoning tasks, respectively. When handling large-scale tables, Chen et al.~\cite{chen2024tablerag} mitigate catastrophic forgetting by introducing a Retrieval-Augmented-Generation (RAG) module and prompt-based reasoning mechanisms to the pipeline. 

While previous research has mainly focused on utilizing LLMs to analyze general structured data~\cite{fang2024large}, approaches for applying LLMs to model domain-specific structured data, such as CT scan protocols, are still lacking. Hence, to alleviate technologists' workload and automate scan protocol modification workflows, this work proposes utilizing an LLM-based agent framework~\cite{xi2025rise} for adaptation of acquisition, reconstruction and postprocessing settings in the device-specific protocol definition files by leveraging the LLM as a planner and providing the agent with specially designed tools.
To systematically evaluate the reliability of the proposed agent framework, a multi-level assessment pipeline with customized metrics has been introduced and illustrated. The agent's performance is compared across different underlying LLMs and among diverse request types. 

\section{Methods}
\subsection{Hierarchical Representation of Scan Protocols}
\begin{figure}[htbp]
    \centering
    \includegraphics[width=0.85\linewidth]{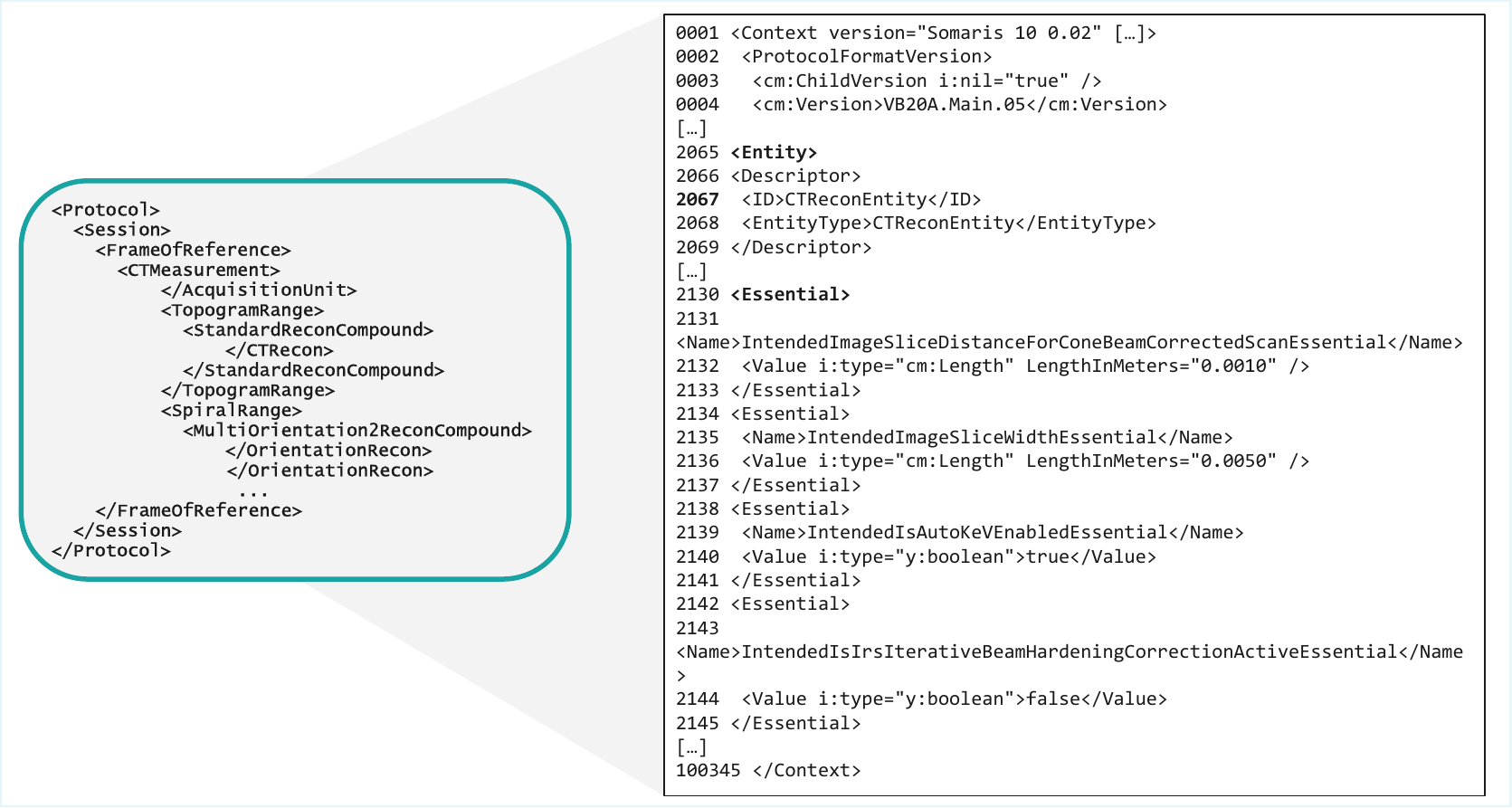}
    \caption{\textbf{XML-based hierarchical structure of a CT scan protocol definition from a commercial scanner.} \textit{Left panel:} exemplary structure and the child-parent relations between different entities. \textit{Right panel:} detailed view of a \texttt{CTRecon} entity with the definition of some essential parameters through key-value pairs.}
    \label{fig:protocol_tree}
\end{figure}
\noindent
CT scan protocols define technical parameters and procedures for examinations. In this work, we investigate vendor-specific CT protocol definition files (Siemens Healthineers, Forchheim, Germany) which are structured as hierarchical XML documents. As shown in Fig.\,\ref{fig:protocol_tree}, these scan protocols adopt a tree structure where each node represents an \textit{\textbf{entity}} corresponding to distinct imaging workflow phases. For instance, \texttt{Range} entities (Topogram/Spiral) define data acquisition phases, while \texttt{Recon} entities (Oriented/Standard) specify reconstruction operations. The hierarchical organization reflects the procedural CT imaging flow, with parent-child relationships indicating the sequence and nesting of scan phases and respective reconstructions. Each entity contains attributes, including \textit{name}, \textit{id}, \textit{type}, and child entities for identification, with critical parameters denoted as \textit{\textbf{essentials}}, controlling acquisition, reconstruction and postprocessing.
\subsection{Agent Framework}
\subsubsection{General Workflow.} As shown in Fig.\,\ref{fig:agent_workflow}, the agent is provided with various tools and a memory component. The process begins with the framework receiving the technologist’s request. Leveraging tool calling, in-context-learning, and reasoning over the fetched structured protocol context, the agent realizes a targeted interaction with the underlying scan protocol and implements the user request with a series of proposed actions. When uncertainty or ambiguity is detected, the agent seeks human feedback to orchestrate the entire workflow, forming a closed-loop enhanced by a \textit{Human-in-the-loop} verification. 
\subsubsection{Router.} 
\begin{figure}[htbp]
    \centering
    \includegraphics[width=0.98\linewidth]{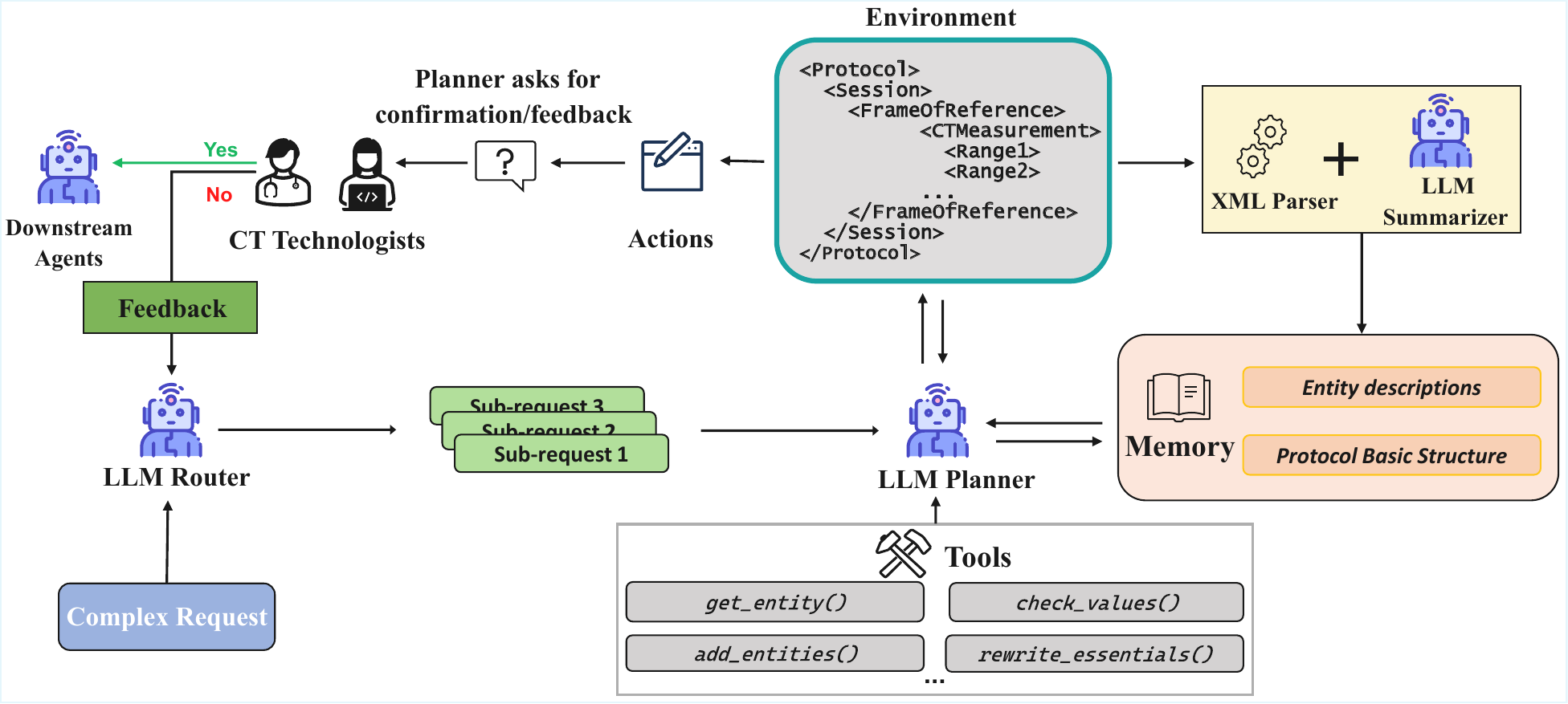}
    \caption{Workflow of the proposed LLM-based agent for scan protocol modification. \textit{\textbf{environment: }}underlying scan protocol, \textit{\textbf{actions: }}modification proposals.}
    \label{fig:agent_workflow}
\end{figure}

Protocol adaptation requests are inherently complex, often requiring multiple operations across diverse components. The Router component addresses this by decomposing user requests into manageable sub-requests and directing them to appropriate downstream agents. Leveraging the LLM's domain knowledge and in-context learning capabilities, the Router classifies sub-requests into four categories: \textit{Adding}, \textit{Modification}, \textit{Deleting}, and \textit{Others}. 
Additionally, few-shot examples are used in prompts to illustrate each category, with the goal of improving classification accuracy. After the query decomposition, except for the \textit{Others} category, the sub-requests are dispatched to their corresponding downstream agents as depicted in Fig.\,\ref{fig:agent_workflow}.  
\subsubsection{Memory.} 
To support consistent decision-making and contextual understanding~\cite{wang2024survey}, we introduce an explicit memory module that stores static and protocol-specific prior knowledge including the following two aspects:\\ 
\emph{Entity Description.} Each protocol entity type is associated with a concise and functional description based on its key configuration values (e.g., slice thickness, kernel type, and reconstruction orientation).\\
\emph{Basic Protocol Structure.} A simplified tree-structured representation of the entire protocol as illustrated in the left hand panel of Fig.\,\ref{fig:protocol_tree} is created by parsing the original XML format protocol.
Leveraging LLM's instruction-following ability~\cite{yuan2024easytool}, the agent makes and then executes plans by dynamically invoking a comprehensive tool set, which includes entity retrieval, attribute management, and validation. 
\subsubsection{Planner.} For each sub-request, the agent retrieves the relevant entities and associated essentials from the XML-scan protocol context. Building upon that and utilizing the LLM's reasoning over structured data and agent's memory section, the planner concludes a concrete proposal to guide the subsequent steps. The planner's output proposal consists of two parts: a structured set of \textit{\textbf{actions}} -- specifying exactly which entities and fields need to be modified -- and a human-readable \textit{\textbf{plan}} that explains these steps clearly. 
\subsubsection{Downstream Agent.} To address diverse requests, the agent first categorizes each request by its type and intent, and then applies tailored processing strategies accordingly. When handling \texttt{Modification} requests, the agent identifies and filters relevant protocol context (\textit{entity}) and then substitutes specific parameter values (\textit{essential}). For requests that involve adding new components, instead of generating new entities from scratch, the agent leverages existing and contextually relevant entities as templates, refines them optionally based on the user intent, and inserts them into appropriate locations in the protocol tree to preserve semantic and hierarchical consistency. In contrast, for deletion-related requests, the agent removes the specified entities and, when an entity is the sole child of its parent, also removes the parent to prevent structural fragmentation so that the modified protocol remains readable by the scanner. 
\subsubsection{Human-In-the-Loop. }To mitigate errors due to LLM limitations, such as hallucinations and ambiguous reasoning, human feedback is incorporated into the framework: after the planner formulates a proposal, it is submitted to the user for review and verification. 
\subsection{Evaluation Methods}
\subsubsection{Syntax Evaluation.} As defined in Eq.\,\ref{eq:scr}, the \textit{Syntax Correctness Rate (SCR)} is deployed to verify that the modified XML protocols have no syntax errors and thus remain fully readable and interpretable by the scanner's protocol browser. 
\begin{equation}
    \label{eq:scr}
    \textnormal{SCR} = \frac{\textnormal{N}_{\textnormal{correct}}}{\textnormal{N}_{\textnormal{total}}},
\end{equation}
where $\textnormal{N}_{\textnormal{correct}}$ is the number of modified protocols that can be successfully compiled by the protocol browser, and $\textnormal{N}_{\textnormal{total}}$ denotes the total number of modified protocols. 
\subsubsection{Semantic Evaluation. }To assess whether the agent accurately interprets the technologist's request and performs faithful adjustments, we further introduce the following measures.
\paragraph{Plan Accuracy:} By comparing the updated XML segments with the ground truth, the accuracy of a request type is computed as: 
\begin{equation}
\textnormal{Plan Accuracy}_{\texttt{t}} = \frac{N_{\textnormal{correct}}}{N_{\textnormal{total}}}, \quad t \in \{\texttt{Modification}, \texttt{Adding}, \texttt{Deleting}\}
\end{equation}
where $N_{correct}$ is the number of requests for which the agent’s modification exactly matches the ground truth, $N_{total}$ is the total number of evaluated requests and $t$ is the request label.
\paragraph{Plan Faithfulness:} We measure plan faithfulness by instructing an LLM to reconstruct pseudo user requests from agent-retrieved entities and essentials, then computing cosine similarity between text embeddings (OpenAI's \textit{text-embedding-003-small}) of the original and inferred tasks:
\begin{equation}
\textnormal{Plan Faithfulness} =\frac{1}{n}\sum_{i=1}^{n}\textnormal{Similarity}(t,,t_i),
\end{equation}
where $t$ and $t_i$ denote embeddings of original and inverse-engineered tasks, respectively, and $n$ is the number of pseudo tasks. 
\subsubsection{Retrieval Evaluation.} We also evaluate whether the agent can fetch expected context from the structured scan protocol on the entity and the essential levels. 
\subsection{Experiment Setup}
In the experiments, we employ an adult thorax CT protocol comprising a topogram, a spiral scan with multi-planar 2D reconstructions using body kernels (Br40) and 3D reconstructions with lung kernels (Bl60). Identical slice settings are applied across multi-planar orientations, followed by LungCAD post-processing. 
In our proposed framework, the agent can process two types of input: (1) natural language requests that describe protocol modifications in free-text form; and (2) structured, JSON-formatted requests, curated to align with the output format of an upstream module in our prototype pipeline. 
Additionally, two API-based LLMs are employed to support the agent: \texttt{GPT-4o}\footnotemark[1] and its lightweight variant \texttt{GPT-4o-mini}\footnotemark[1]. Additionally, to evaluate deployment scenarios in which cost efficiency and data privacy are prioritized higher, the evaluation also includes a locally deployed open-source LLM (\texttt{Gemma3:27B}\footnotemark[2]), running on an HPC cluster with 8 NVIDIA V100 32GB GPUs. Notably, all of the deployed LLMs are used without any task-specific fine-tuning. 
\footnotetext[1]{API version: \texttt{2024-10-28}}
\footnotetext[2]{Maximum input tokens: $64{,}000$}
\section{Results}
\subsection{Quantitative Evaluation}
\begin{figure}[!htbp]
    \centering
    \includegraphics[width=0.75\linewidth]{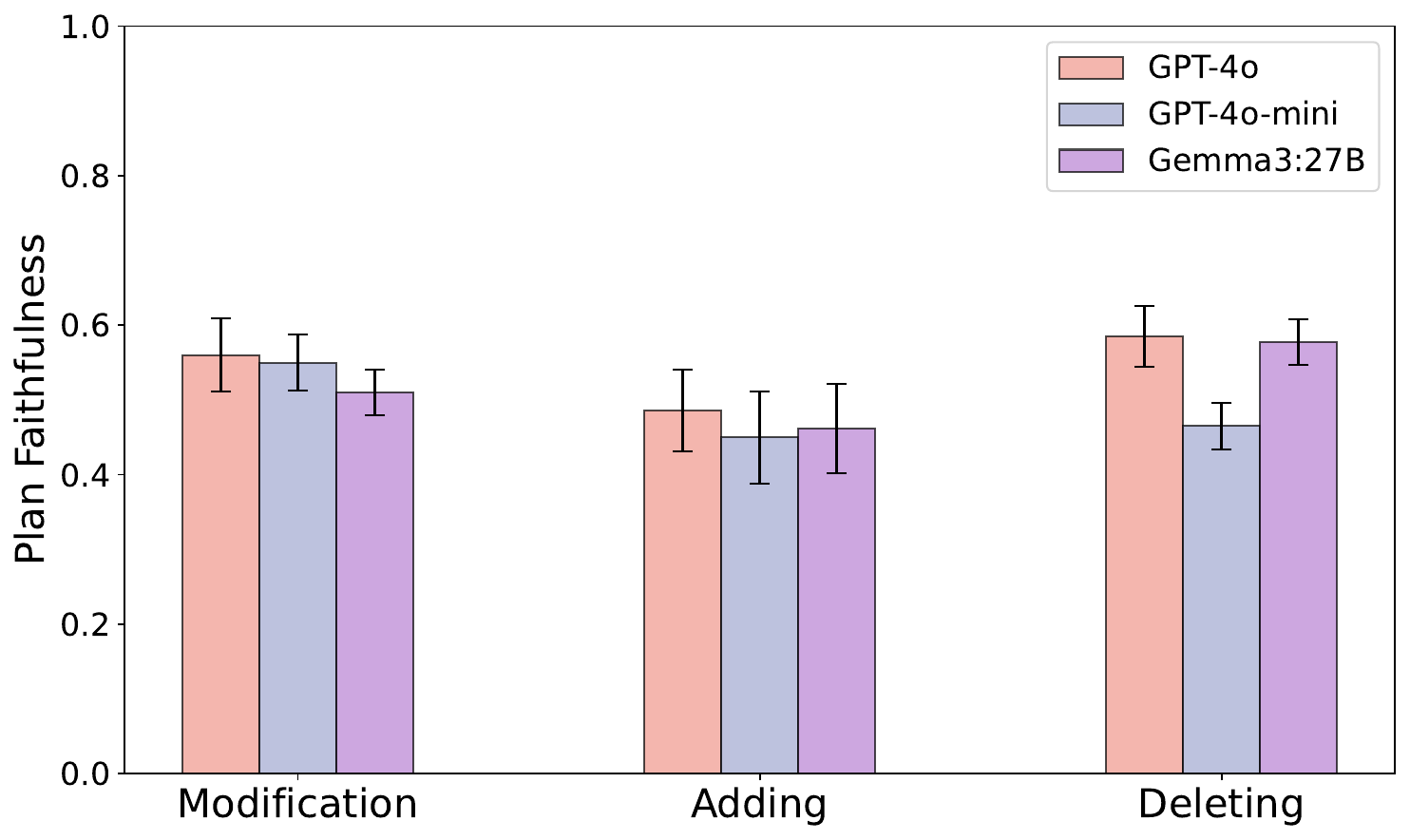}
    \caption{\textit{Plan Faithfulness} of the agent with different LLMs. Error bars indicate the standard error of the mean. \texttt{GPT-4o} is utilized to generate \textbf{10} pseudo-tasks from the agent-retrieved-protocol contexts for each real request. }
    \label{fig:sim_score} 
\end{figure}

\begin{figure}
    \centering
    \begin{subfigure}[t]{0.49\textwidth}
        \centering
        \includegraphics[width=\linewidth]{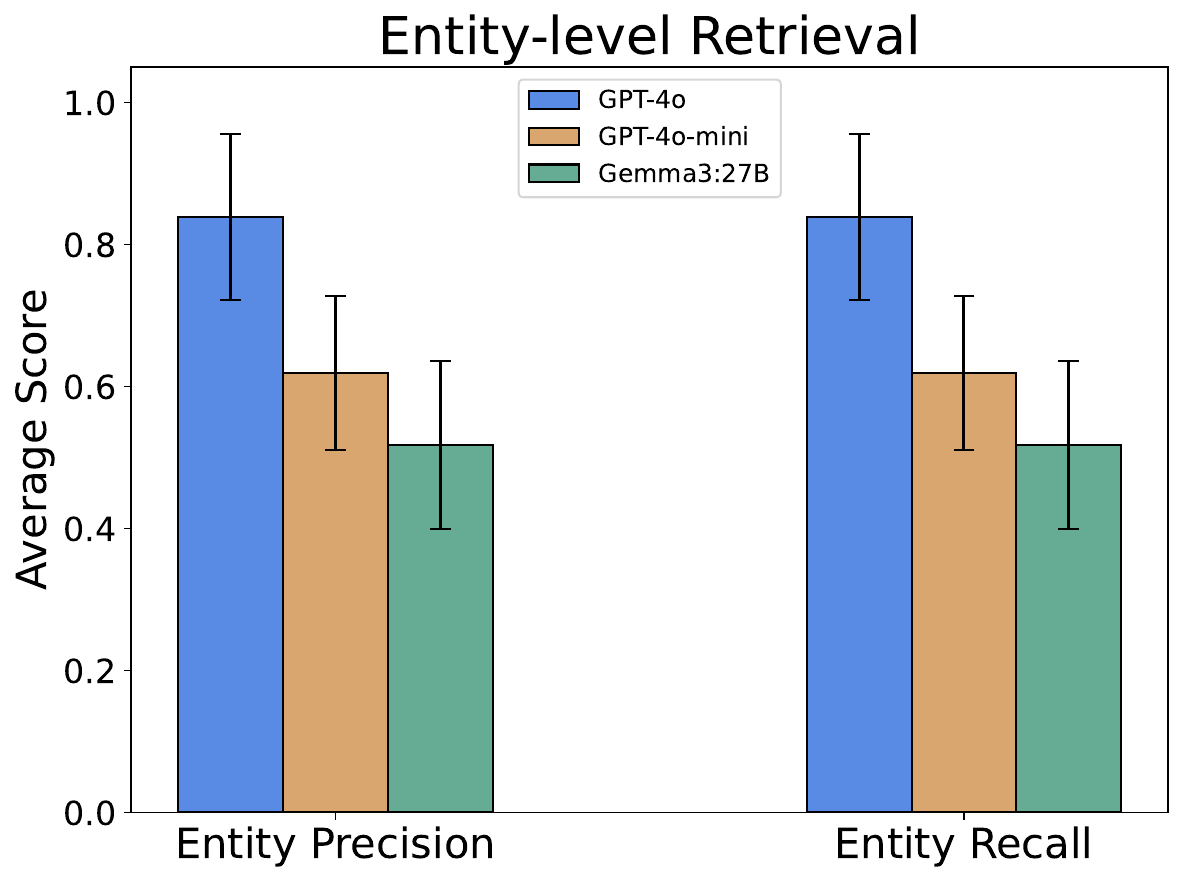}
        \caption*{\textbf{(a)}}
    \end{subfigure}
    \hfill
    \begin{subfigure}[t]{0.49\textwidth}
        \centering
        \includegraphics[width=\linewidth]{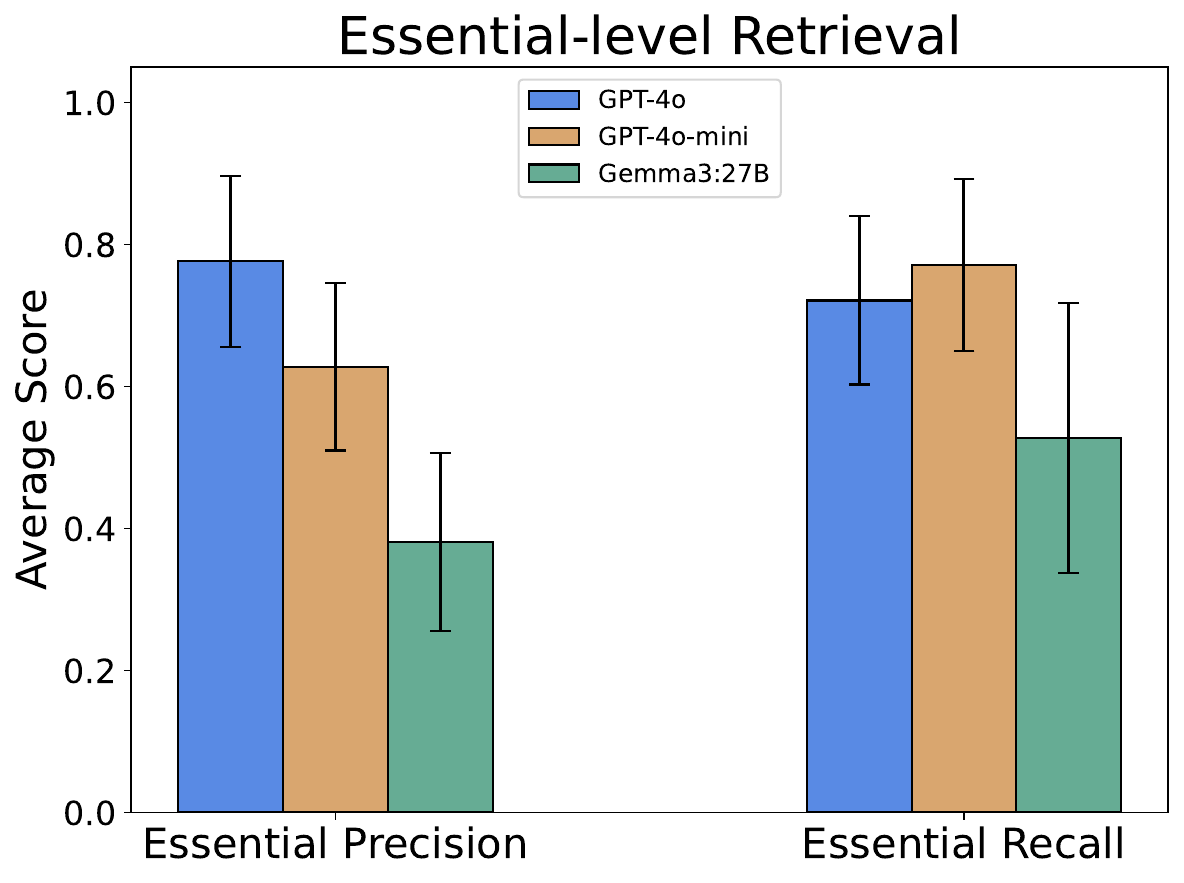}
        \caption*{\textbf{(b)}}
    \end{subfigure}
    \caption{Agent retrieval performance across different LLMs. Error bars indicate the standard error of the mean. }
    \label{fig:retrieval_scores}
\end{figure}

Tab.\,\ref{tab:exp_scr} presents a comparative analysis of the agent's XML syntax-preserving capabilities across LLMs. In general, \texttt{GPT-4o} achieves the highest overall SCR ($83\%$), particularly excelling in structure-sensitive tasks like \textit{Adding} and \textit{Deleting}. While \texttt{GPT-4o-mini} performs on par with \texttt{GPT-4o} in \textit{Deleting} tasks, it under-performs in other categories. Despite a lower general SCR score, \texttt{Gemma3:27B} surpasses \texttt{GPT-4o-mini} in handling \textit{Adding} tasks.
The semantic evaluation results in Tab.~\ref{tab:accuracy} and Fig.~\ref{fig:sim_score} further confirm the consistent advantage of \texttt{GPT-4o}. \texttt{GPT-4o-mini} shows reduced faithfulness on complex requests, while \texttt{Gemma3:27B} demonstrates selective strengths in \textit{Deleting} and structured input interpretation. Furthermore, Fig.~\ref{fig:retrieval_scores} illustrates the agent's retrieval performance, where \texttt{GPT-4o} achieves the best performance in retrieving entity-level protocol context. 
Beyond the quantitative analysis, some qualitative results can be found in the supplementary material, demonstrating the agent's decision making process during implementing the user request. 
\begin{table}[!htbp]
    \centering
    \caption{\textbf{SCR of the agents with different LLMs.} Here, the category of \textbf{JSON} refers to the structured system inputs, which are curated to align with the output format of an upstream module in the prototype pipeline. The general score is the arithmetic average over the four categories.}
    \vspace{0.5em}
    \begin{tabular}{cccccc}
         \toprule
         \multirow{2}*{\textbf{Model}} & \multicolumn{4}{c}{\textbf{Request Types}} & \multirow{2}*{\textbf{\textit{General}}}\\
         \cmidrule(lr){2-5}& \textbf{\texttt{Modification}} & \texttt{\textbf{Adding}} & \texttt{\textbf{Deleting}} & \textbf{JSON} & \\
         \midrule
         \texttt{GPT-4o} & \textbf{0.89} & \textbf{0.80} & \textbf{0.75} & \textbf{0.80} & \textbf{0.83} \\
         \texttt{GPT-4o-mini} & 0.67 & 0.20 & \textbf{0.75} & 0.60 & 0.57 \\
         \texttt{Gemma3:27B\footnotemark} & 0.56 & 0.40 & 0.50 & 0.60 & 0.43 \\
         \bottomrule
    \end{tabular}
    \label{tab:exp_scr}
\end{table}
\begin{table}[!htbp]
        \centering
        \caption{\textit{Plan Accuracy} of the agents with different LLMs. }
        \vspace{0.5em}
        \begin{tabular}{@{}lccccc@{}}
        \toprule
            \multirow{2}*{\textbf{Model}} & \multicolumn{4}{c}{\textbf{Request Type}} & \multirow{2}*{\textbf{\textit{General}}} \\
            \cmidrule(lr){2-5} & \texttt{\textbf{Modification}} & \texttt{\textbf{Adding}} & \textbf{\texttt{Deleting}} & \textbf{JSON} \\
            \midrule
             \texttt{GPT-4o} & \textbf{0.67} & \textbf{0.40} & 0.25 & \textbf{0.60} & \textbf{0.52}\\
             \texttt{GPT-4o-mini} & 0.33 & 0.20 & 0.25 & 0.20 & 0.26\\
             \texttt{Gemma3:27B} & 0.33 & 0.20 & \textbf{0.50} & \textbf{0.60} & 0.39\\
             \bottomrule
        \end{tabular}
        \label{tab:accuracy}
\end{table}
\subsection{Error Analysis}
To highlight the agent's limitations and guide further improvement, it is important to assess the common failure cases that occurred during the agent's attempt to modify the CT scan protocols. A typical reason for errors is misclassification by the router, leading to the dispatch of the task to a wrong downstream agent. 
Furthermore, we encounter the structural errors that result in dangling or broken structures, where the agent modifies the XML hierarchy incorrectly. For instance, the agent could add a a new entity under the wrong parent node or delete a sole child node without removing its parent. 
Additionally, due to the agent's lack of device-specific knowledge, it may fail to understand the inter-dependencies between attributes and use values that are not supported in the given context, which leads to a semantically incorrect protocol modification. 
\section{Discussion and Conclusion}
This study explores how LLM agents modify device-specific XML-based CT scan protocols from natural or structured clinical inputs, leveraging their reasoning and domain comprehension.
Building upon this, to comprehensively assess the effectiveness and robustness of the proposed approach, a multi-aspect evaluation pipeline was designed. Results demonstrate that with appropriate LLMs, the agent can already achieve considerable success in generating syntactically correct and user-intent-aligned scan protocols.

Despite these promising outcomes, our study still faces challenges. While we aim to mitigate the data-scarcity issue by generating complementary pseudo-requests using LLMs and text embedding models, this potentially introduces bias due to inconsistent model performance across different scenarios.
Another limitation relates to the lack of comprehensive device-internal knowledge. The proposed framework provides the agents with scan protocol context and background knowledge through input prompts. Nonetheless, the agents have an insufficient understanding of attribute inter-dependencies within protocols, device-specific naming conventions for acquisition and reconstruction parameters, and supported value ranges for protocol variables, which in turn may dynamically depend on the current values of other parameters. This knowledge gap may mislead the agent into incorrect reasoning and thus result in erroneous modifications.

To this end, future research should focus on expanding the agent's access to relevant knowledge sources, in particular semantically meaningful device APIs, but also structured radiology knowledge bases. Developing more robust evaluation methodologies that minimize model-dependent biases would strengthen the assessment of agent performance and reliability in real-world clinical settings. 

In conclusion, as a proof of concept, our work demonstrates the feasibility of using LLM agents for retrieving information from real protocol definition files and generating valid, device-compatible modifications for them based on various clinical requests. In addition, it explored how to systematically assess LLM agents' capabilities in a complex and domain-specific setting.
\begin{credits}
\subsubsection{\discintname}
The authors have no competing interests to declare that are relevant to the content of this article.
\end{credits}

\section*{Supplementary Materials}
\titlerunning{Towards Autonomous CT Protocol Management}
\begin{tcolorbox}[markdownbox, title=Protocol Adjustment Log, colback=blue!5]
    \textbf{User Request:} \\
    \texttt{we should do a lateral topo instead of the AP. You can put the tube position to right or left}\\
    \textbf{Deployment:} \texttt{gpt-4o-mini}\\
    \textbf{Router:}
    \begin{itemize}
      \item \texttt{User Intention:}modify
      \item \texttt{Classification Reason: }The request suggests changing from the AP method to a lateral topo and adjusting the tube position, which indicates a modification of the original approach.
    \end{itemize}
    \vspace{2mm}
    \textbf{Modification Plan:}
    \begin{enumerate}
        \item Change the topogram type in the topogram measurement range to lateral orientation by updating the table direction from 'HeadToFeet' to 'Lateral'. This adjustment is crucial within the topogram measurement period of the protocol\footnotemark[1].
        \item Adjust the tube position in the acquisition unit for the CT measurement by specifying it as either 'Right' or 'Left', based on the user's request. This should be reflected in the acquisition unit settings.
    \end{enumerate}
    \vspace{2mm}
    \textbf{Retrieved Relevant Entities and Essentials:}
    (please see next page)
\end{tcolorbox}
\newpage
\footnotetext[1]{Due to the LLM agent's lack of device-specific knowledge, it plans an incorrect modification of the table direction to perform a lateral topogram scan. }
\begin{tcolorbox}[markdownbox, title=Protocol Adjustment Log (continued), colback=blue!5]
\begin{itemize}
    \item \textbf{Modification Parameter: }\texttt{Topogram Type}
    
    \textbf{Suggested Modification:} Replace the anterior-posterior (AP) topogram with a lateral topogram.
    
    \textbf{Relevant Protocol Elements: }
    \begin{itemize}
        \item \textbf{Target Entity: }\texttt{TopogramRangeEntity (ID:...)}

        \textbf{Selected Essential(s): }
        \begin{itemize}
            \item \textbf{Name: }\texttt{TableDirectionPatientRelatedEssential}
            \item \textbf{In XML format: }
        \end{itemize}
    \end{itemize}
\begin{lstlisting}[basicstyle=\scriptsize, backgroundcolor=\color{gray!10}, frame=single]
<Essential>
  <Name>TableDirectionPatientRelatedEssential</Name>
  <Value i:type="TableDirectionPatientRelated">Lateral</Value>
</Essential>
        \end{lstlisting}
    \item \textbf{Modification Parameter: }\texttt{Tube Position}
    
    \textbf{Suggested Modification:} Set the tube position to either right or left.
    
    \textbf{Relevant Protocol Elements: }
    \begin{itemize}
        \item \textbf{Target Entity: }\texttt{AcquisitionUnitEntity (ID:...)}
        \textbf{Selected Essential(s): }
        \begin{itemize}
            \item \textbf{Name: }\texttt{PerformedTubeCurrentProfileEssential}
            \item \textbf{In XML format: }
        \end{itemize}
    \end{itemize}
\begin{lstlisting}[basicstyle=\scriptsize, backgroundcolor=\color{gray!10}, frame=single]
<Essential>
    <Name>PerformedTubeCurrentProfileEssential</Name>
    <Value i:type="SpatialCurrentProfile">
      <PositionsWithCurrents>
         <Position>Right</Position>
      </PositionsWithCurrents>
    </Value>
</Essential>
\end{lstlisting}
\end{itemize}
\end{tcolorbox}

\begin{tcolorbox}[markdownbox, title=Protocol Adjustment Log, colback=blue!5]
\textbf{User Request:} \\
\texttt{for this protocol, the patient should be positioned face up and feet first}

\vspace{2mm}

\textbf{Deployment:} \texttt{gpt-4o-mini}

\vspace{2mm}

\textbf{Router:}
\begin{itemize}
  \item \texttt{User Intention:}modify
  \item \texttt{Classification Reason:} The request indicates a specific instruction about positioning a patient, which suggests that there may be a need to adjust or modify existing protocols regarding patient positioning.
\end{itemize}

\vspace{2mm}

\textbf{Modification Plan:}
To modify the patient positioning in this protocol, please adjust the positioning of the patient to be face up and feet first. This modification specifically affects the essential patient positioning attribute within the frame of reference section of the protocol. Change the patient positioning value from 'face up head first' to 'face up feet first' to ensure accurate alignment for the imaging procedure.

\vspace{2mm}

\textbf{Retrieved Relevant Entities and Essentials:}
\begin{itemize}
    \item \textbf{Modification Parameter:} \texttt{Patient Positioning}
    
    \textbf{Suggested Modification:} \texttt{Position the patient face up and feet first for this protocol.}
    
    \textbf{Relevant Protocol Elements:}
    \begin{itemize}
  \item \textbf{Target entity:} \texttt{FrameOfReferenceEntity}
  \item \textbf{Selected essential(s):}
  \begin{itemize}
    \item Name: \texttt{PatientPositionEssential}
    \item In XML format:
  \end{itemize}
\end{itemize}
\end{itemize}
\begin{lstlisting}[basicstyle=\scriptsize, backgroundcolor=\color{gray!10}, frame=single]
<Essential>
  <Name>PatientPositionEssential</Name>
  <Value i:type="cm:PatientPosition">FaceUpFeetFirst</Value>
</Essential>
    \end{lstlisting}
\end{tcolorbox}

\begin{tcolorbox}[markdownbox, title=Protocol Adjustment Log, colback=blue!5]

\textbf{User Request:} \\
\texttt{can you delete the lung cad from this protocol?}

\vspace{2mm}

\textbf{Deployment:} \texttt{gpt-4o-mini}

\vspace{2mm}

\textbf{Router:}
\begin{itemize}
  \item \texttt{User Intention: delete}
  \item \texttt{Classification Reason: The request specifically asks to remove an item (lung cad) from a protocol, which corresponds to the action type 'delete'.}
\end{itemize}

\vspace{2mm}

\textbf{LLM Generated Plan:}
Remove the CT recon entity associated with the inline result for LungCAD, as it is no longer necessary for the protocol. This will also entail the removal of the standard reconstruction compound that contains this recon entity since it is linked to the inline result that is being eliminated. This cleaning of the protocol will streamline the scanning process and eliminate outdated or unnecessary data.
\vspace{2mm}

\textbf{Entities to be Removed:}
\begin{itemize}
    \item \textbf{Target Entity 1}
    \begin{itemize}
      \item \textbf{Entity Name:} \texttt{CTReconEntity}
      \item \textbf{Entity ID:} \texttt{...}
      \item \textbf{Reason:} The entity 'Inline Result: LungCAD' is associated with this CTReconEntity and needs to be removed.
    \end{itemize}
    \item \textbf{Target Entity 2}
        \begin{itemize}
          \item \textbf{Entity Name:} \texttt{StandardReconCompoundEntity}
          \item \textbf{Entity ID:} \texttt{...}
          \item \textbf{Reason:} The CTReconEntity is part of this StandardReconCompoundEntity; thus, it needs to be removed as well.
        \end{itemize}

\end{itemize}
\vspace{1mm}

\end{tcolorbox}
%
%
%
\bibliographystyle{splncs04}
\bibliography{scan-do}

\begin{thebibliography}{1}
\providecommand{\url}[1]{\texttt{#1}}
\providecommand{\urlprefix}{URL }
\providecommand{\doi}[1]{https://doi.org/#1}

\bibitem{alexander2022mandating}
Alexander, R., Waite, S., Bruno, M.A., Krupinski, E.A., Berlin, L., Macknik, S., Martinez-Conde, S.: Mandating limits on workload, duty, and speed in radiology. Radiology  \textbf{304}(2),  274--282 (2022)

\bibitem{chen2024tablerag}
Chen, S.A., Miculicich, L., Eisenschlos, J., Wang, Z., Wang, Z., Chen, Y., Fujii, Y., Lin, H.T., Lee, C.Y., Pfister, T.: Tablerag: Million-token table understanding with language models. Advances in Neural Information Processing Systems  \textbf{37},  74899--74921 (2024)

\bibitem{d2024large}
D’Antonoli, T.A., Stanzione, A., Bluethgen, C., Vernuccio, F., Ugga, L., Klontzas, M.E., Cuocolo, R., Cannella, R., Ko{\c{c}}ak, B.: Large language models in radiology: fundamentals, applications, ethical considerations, risks, and future directions. Diagnostic and Interventional Radiology  \textbf{30}(2), ~80 (2024)

\bibitem{fang2024large}
Fang, X., Xu, W., Tan, F.A., Zhang, J., Hu, Z., Qi, Y., Nickleach, S., Socolinsky, D., Sengamedu, S., Faloutsos, C.: Large language models (llms) on tabular data: Prediction, generation, and understanding--a survey. arXiv preprint arXiv:2402.17944  (2024)

\bibitem{hegselmann2023tabllm}
Hegselmann, S., Buendia, A., Lang, H., Agrawal, M., Jiang, X., Sontag, D.: Tabllm: Few-shot classification of tabular data with large language models. In: International Conference on Artificial Intelligence and Statistics. pp. 5549--5581. PMLR (2023)

\bibitem{jiang2023structgpt}
Jiang, J., Zhou, K., Dong, Z., Ye, K., Zhao, W.X., Wen, J.R.: Structgpt: A general framework for large language model to reason over structured data. arXiv preprint arXiv:2305.09645  (2023)

\bibitem{wang2024survey}
Wang, L., Ma, C., Feng, X., Zhang, Z., Yang, H., Zhang, J., Chen, Z., Tang, J., Chen, X., Lin, Y., et~al.: A survey on large language model based autonomous agents. Frontiers of Computer Science  \textbf{18}(6),  186345 (2024)

\bibitem{xi2025rise}
Xi, Z., Chen, W., Guo, X., He, W., Ding, Y., Hong, B., Zhang, M., Wang, J., Jin, S., Zhou, E., et~al.: The rise and potential of large language model based agents: A survey. Science China Information Sciences  \textbf{68}(2),  121101 (2025)

\bibitem{yuan2024easytool}
Yuan, S., Song, K., Chen, J., Tan, X., Shen, Y., Kan, R., Li, D., Yang, D.: Easytool: Enhancing llm-based agents with concise tool instruction. arXiv preprint arXiv:2401.06201  (2024)

\end{thebibliography}

\end{document}